\pdfoutput=1
\documentclass[11pt]{article}

\usepackage{EMNLP2023}

\usepackage{times}
\usepackage{latexsym}

\usepackage[T1]{fontenc}
\usepackage[utf8]{inputenc}

\usepackage{microtype}

\usepackage{inconsolata}

\usepackage{times}
\usepackage{latexsym}
\usepackage{soul}
\usepackage{booktabs}
\usepackage{amsmath,caption}
\usepackage{colortbl}
\usepackage{arydshln}
\usepackage{multirow}
\usepackage{multicol}

\usepackage{microtype}
\usepackage{graphicx} 
\usepackage{caption,subcaption}

\title{
Batch Prompting: Efficient Inference with Large Language Model APIs
}

\author{Zhoujun Cheng \\
  Shanghai Jiao Tong University \\
  \texttt{blankcheng@sjtu.edu.cn} \\\And
  Jungo Kasai \\
  University of Washington \\
  \texttt{jkasai@cs.washington.edu} \\\And
  Tao Yu \\
  University of Hong Kong\\
  \texttt{tyu@cs.hku.hk}\\
}
\begin{document}
\maketitle
\begin{abstract}
Performing inference on large volumes of samples with large language models (LLMs) can be computationally and financially costly in industry and real-world use.
We propose batch prompting, a simple yet effective prompting approach that enables the LLM to run inference in batches, instead of one sample at a time.
Our method reduces both token and time costs while retaining downstream performance.
We theoretically demonstrate that under a few-shot in-context learning setting, the inference costs decrease almost inverse linearly with the number of samples in each batch.
We extensively validate the effectiveness of batch prompting on ten datasets across commonsense QA, arithmetic reasoning, and NLI/NLU: batch prompting significantly~(up to $5\times$ with six samples in batch) reduces the LLM (Codex) inference token and time costs while achieving better or comparable performance. 
For state-of-the-art Chat-based LLMs, e.g., GPT-3.5 and GPT-4, we show the benefits of batch prompting also hold.
Further analysis shows that the number of samples in each batch and the complexity of tasks affect its performance.
Moreover, batch prompting can be applied across different reasoning methods using LLMs. Our code can be found at the site \url{https://github.com/xlang-ai/batch-prompting}.
\end{abstract}

% ----------------------------------------------------------------------------------------
% ----------------------------------------------------------------------------------------
\section{Introduction}
\label{sec:introdeuction}

\begin{figure}[t]
    \begin{center}
    \includegraphics[width=2.7in]{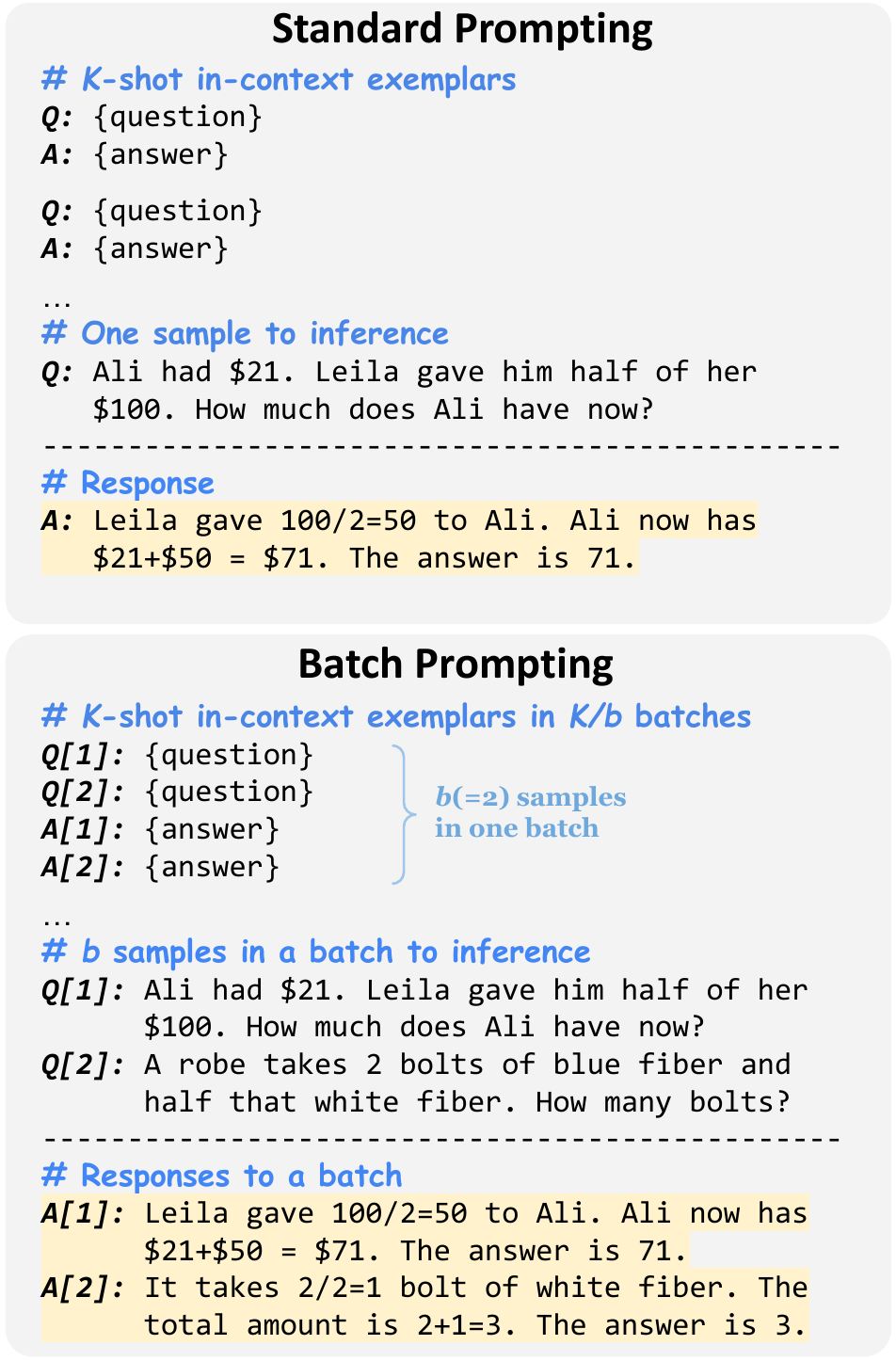}
    \end{center}
\caption{Illustration of batch prompting compared with standard prompting. Batch prompting groups multiple samples in one batch ($b\!=\!2$ in the figure) and lets the LLM generate multiple responses (highlighted in \textcolor{yellow}{yellow}) for the batch in inference.}
\label{fig:overview}
\vspace{-0.35cm}
\end{figure}

Large language models~(LLMs) have shown their strong capabilities under zero/few-shot settings with in-context learning~\cite{gpt3,codex,palm,ouyang2022training}. 
Much recent work has made progress in in-context learning by eliciting reasoning steps~\cite{cot,wang2022self,khot2022decomposed,cheng2022binding, yao2022react}, selecting representative in-context exemplars~\cite{liu2022makes,su2022selective,Agrawal2022IncontextES}, and designing prompt templates~\cite{jiang2020can,bach2022promptsource,arora2022ask}.

Using LLMs can be costly in terms of token and time usage, especially when large volumes of LLM calls are needed, such as benchmarking a large dataset or addressing a high volume of customer inquiries for businesses.
For example, the widely-adopted OpenAI API service\footnote{\url{https://openai.com/api/}.} of LLMs
requires about $\$40$ and $8$ hours to perform inference on $10$K samples using gpt-3.5-turbo; and the expense significantly escalates when using gpt-4, exceeding a substantial $\$600$. \footnote{Assume each LLM call consumes $2,000$ tokens~(common for few-shot or long instruction), including both the input prompt tokens and generated tokens, and each call takes $3$ seconds to finish~(a plausible average time in real use).}
If the rate limits of maximum API requests per minute are also considered, the costs will be even higher, preventing users from building massive LLM applications. 

We propose batch prompting, a simple yet effective approach for prompting LLMs, which allows the model to perform inference on multiple samples at once, instead of one sample at a time.
This reduces token and time costs while still retaining downstream performance, \emph{without} any change in APIs.
As shown in Figure~\ref{fig:overview}, standard prompting generates a response (answer) to one sample at a time, which takes $N$ inference runs of an LLM for a test set of size $N$.
For our batch prompting, on the other hand, an LLM generates responses to $b$ samples in a single inference run and only takes $N/b$ runs for the same $N$ samples.

We first demonstrate theoretically that under the few-shot in-context learning setting, most tokens consumed during the API call are the few-shot exemplars, and only a small portion of token budgets are used for the particular inference sample(s) (Section~\ref{sec:approach}). 
Therefore, increasing $b$ in batch prompting reduces the token and time costs in an inverse linear fashion.
We extensively validate the effectiveness of batch prompting on diverse 
% \jungo{check}
downstream datasets across commonsense QA, arithmetics, and NLI/NLU using Codex, a strong variant of GPT-3 finetuned on code data (Section~\ref{sec:experiments}).
We also test batch prompting on the state-of-the-art GPT-3.5 and GPT-4 models.
Batch prompting significantly decreases the tokens and run time of using LLMs while achieving comparable or even better performance on all ten datasets.

%with Codex~(code-davinci-002).
In further analysis~(Section~\ref{sec:analysis}), we find the number of samples in batch and the complexity of tasks affect its performance.
Moreover, we show that batch prompting works well across different reasoning methods~(\textit{e.g.}, end-to-end, Chain-of-Thought, and code generation), suggesting that batch prompting is an efficient drop-in substitute for conventional prompting.

% ----------------------------------------------------------------------------------------
% ----------------------------------------------------------------------------------------
\section{Approach}
\label{sec:approach}

\begin{figure*}[htb]
    \centering % <-- added
\begin{subfigure}{0.32\textwidth}
  \includegraphics[width=\linewidth]{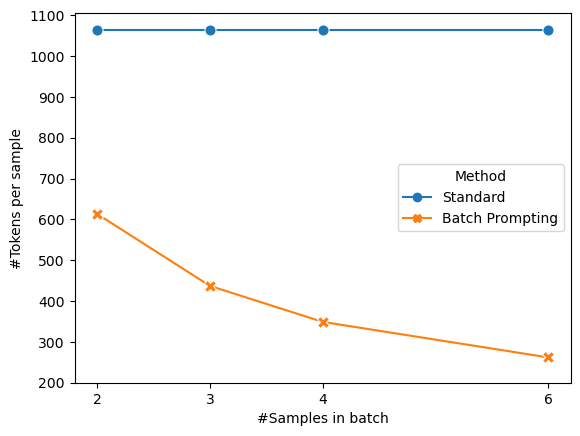}
  \caption{Token(CommonsenseQA)}
  \label{fig:1}
\end{subfigure}\hfil % <-- added
\begin{subfigure}{0.32\textwidth}
  \includegraphics[width=\linewidth]{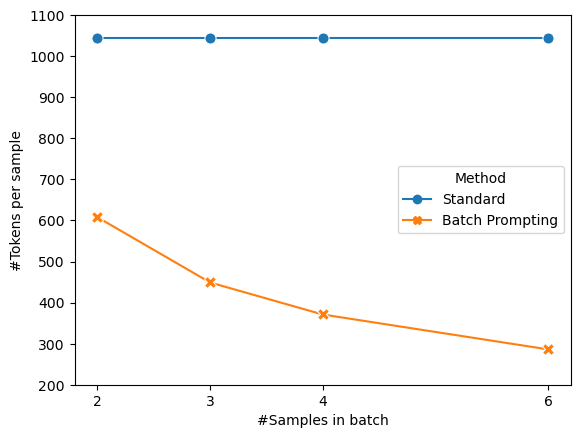}
  \caption{Token(GSM8K)}
  \label{fig:2}
\end{subfigure}\hfil % <-- added
\begin{subfigure}{0.32\textwidth}
  \includegraphics[width=\linewidth]{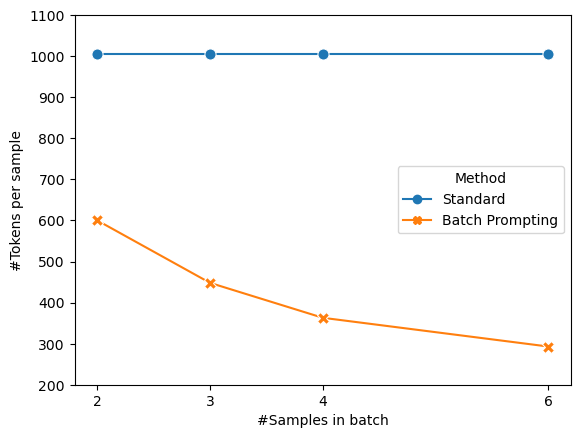}
  \caption{Token(RTE)}
  \label{fig:3}
\end{subfigure}

\medskip
\begin{subfigure}{0.32\textwidth}
  \includegraphics[width=\linewidth]{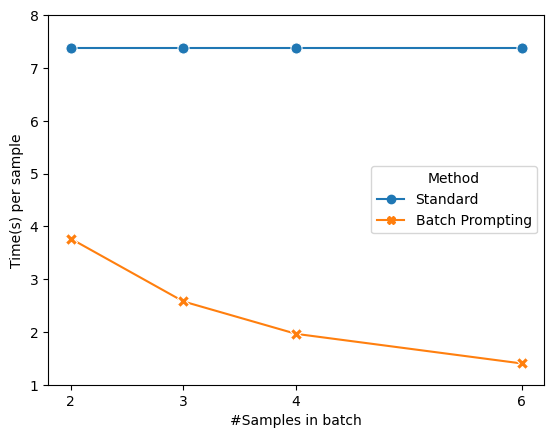}
  \caption{Time(CommonsenseQA)}
  \label{fig:4}
\end{subfigure}\hfil % <-- added
\begin{subfigure}{0.32\textwidth}
  \includegraphics[width=\linewidth]{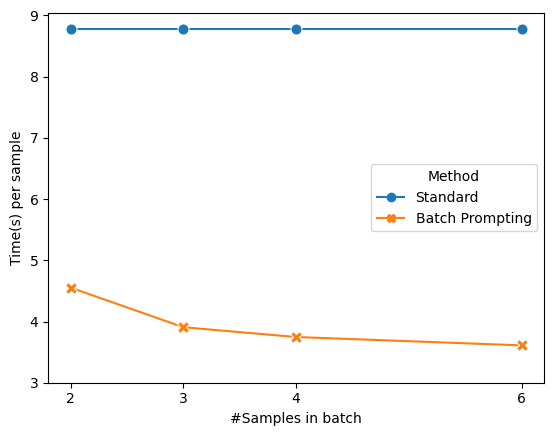}
  \caption{Time(GSM8K)}
  \label{fig:5}
\end{subfigure}\hfil % <-- added
\begin{subfigure}{0.32\textwidth}
  \includegraphics[width=\linewidth]{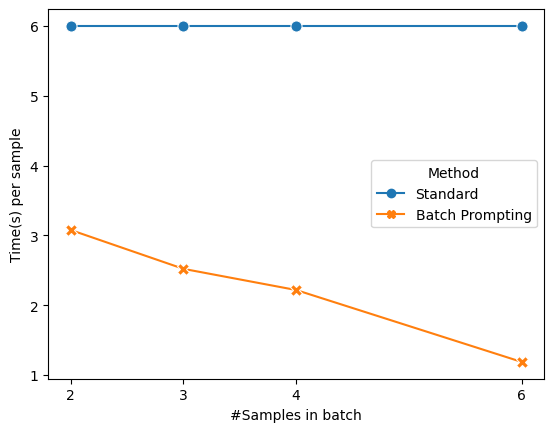}
  \caption{Time(RTE)}
  \label{fig:6}
\end{subfigure}
\caption{Token and time costs per sample on three datasets for illustrations~(other datasets show similar trends). Batch prompting significantly lowers both token and time costs as the number of samples in each batch increases.}
\label{fig:main cost}
\vspace{-0.3cm}
\end{figure*}

We first introduce batch prompting, an efficient alternative to standard prompting.
We then compare the token and time costs of batch and standard prompting, demonstrating the efficiency of our method.

% ------------------------------------------------------------
\subsection{Problem Setup}
The conventional paradigm~(\textit{i.e.}, standard prompting in Figure~\ref{fig:overview}) to prompt LLMs for in-context learning is as follows:  $K$ in-context few-shot exemplars with both a context~(\textit{e.g.}, question) and an output~(\textit{e.g.}, answer) are selected to build the input prompt, \emph{one} test sample with context only is appended at the end of the prompt, and the LLM is used to generate the response for the test sample.

In this paper, we focus on a realistic scenario with $N$ test samples in total, which is common when benchmarking on a dataset or handling a large volume of customer requests.
In this case, it takes $N$ separate calls of the LLM inference under the standard prompting paradigm.

% ------------------------------------------------------------
\subsection{Batch Prompting}
Batch prompting enables the LLM to generate responses for multiple samples in one batch in a \emph{single} inference run, so that it reduces the LLM inference time from $N$ to $N/b$, where $b$ is the number of samples in one batch.
Specifically, as shown in Figure~\ref{fig:overview}, our prompt groups the $K$ in-context exemplars into $K/b$ batches with $b$ exemplars each as demonstrations. 
In every batch, demonstration contexts are arranged in a specific order at the beginning, with their corresponding outputs placed in the same order afterwards. 
Then, $b$ test sample contexts are grouped together at the end of the input prompt. 
In this way, the LLM learns from the in-context demonstrations and generates corresponding responses for the entire batch of test samples.
We add a position identifier ``$\left[index\right]$'' within each batch to 1) assist the LLM with identifying the order correspondence of input contexts and generated responses and 2) ease the process of parsing the generated responses.
% ------------------------------------------------------------
\subsection{Token Cost}
\label{subsec:token cost}
The costs of one LLM call scale linearly with the number of \textit{tokens}, including both the input \textit{prompt tokens} (few-shot and instruction) and \textit{generated tokens}
~(according to, for example, OpenAI's pricing). 
\textbf{Most tokens are consumed by the prompt tokens in standard prompting} because the number of prompt tokens is usually far more than the number of generated tokens so that the LLM can better learn from in-context exemplar. 
Thus, the larger the portion of tokens spent on generated tokens, the more economical the total cost is. 

We define \textit{token efficiency} $\eta$ as the portion of tokens spent on generated tokens in one LLM call. 
For standard prompting and batch prompting~(the instruction tokens are omitted if any for brevity):

\begin{equation}
    \begin{aligned}
        \eta_{standard} = \frac{1}{K + 1} \\
        \eta_{batch} = \frac{b}{K + b}
    \end{aligned}
\end{equation}

When $K \gg 1$ and $b < K$, $\eta_{batch}$ scales almost inverse linearly with $b$, and thus increasing $b$ of batch prompting can greatly reduce token costs.

% ------------------------------------------------------------
\subsection{Time Cost}
\label{subsec:time cost}
Intuitively, batch prompting reduces the inference time by decreasing the number of API calls from $N$ to $N/b$. 
Considering the Transformer \cite{vaswani2017attention} decoding time, the cost will increase with $b$ in batch prompting due to the generation of longer responses compared to standard prompting.
We give a detailed derivation from Transformer architecture perspective in Appendix~\ref{appendix:time analysis}. 

However, as most end-users are accustomed to and only have access to LLM API services, this part of time cost is marginal~(observed in main experiments), relative to the overhead of API call and request rate limits per minute set by a company, such as OpenAI. 
Besides, cases may occur when network connections are unstable or slow, and the users seek to finish a task with as few LLM calls as possible.

Therefore, in practice, reducing the number of calls from $N$ to $N/b$ with batch prompting can essentially lower the time costs.
Note that when the API call overhead and rate limits are no longer the major bottlenecks of time costs in the future, then the increased decoding time to generate longer sequences discussed in Appendix~\ref{appendix:time analysis} cannot be overlooked, and the time reduction of batch prompting will not be as pronounced.

Since LLM infrastructure/services can change over time, the token cost comparison is more reliable and durable to measure than time costs.

% ----------------------------------------------------------------------------------------
% ----------------------------------------------------------------------------------------
\section{Experiments}
\label{sec:experiments}
We extensively evaluate batch prompting across ten diverse datasets. Our results suggest that batch prompting can achieve at most $5\times$ token and time efficiency~(with six samples in batches) improvement with similar or even better downstream performance.
% ------------------------------------------------------------
\subsection{Datasets}
We evaluate batch prompting on ten datasets across commonsense question answering, arithmetic reasoning, and natural language understanding/inference: CommonsenseQA~\cite{commonsenseqa}, StrategyQA~\cite{strategyqa}, GSM8K~\cite{gsm8k}, SVAMP~\cite{svamp}, AQuA~\cite{aqua}, AddSub~\cite{addsub}, MultiArith~\cite{multiarith}, RTE~\cite{rte}, MNLI~\cite{mnli}, and SST-5~\cite{sst-5}.
For CommonsenseQA, AQuA, AddSub, MultiArith, and RTE, we evaluate the whole dev/test sets. For the other five datasets, we evaluate the first $300$ test samples considering the costs of LLM APIs.

% ------------------------------------------------------------
\subsection{Experimental Setups}
We evaluate OpenAI Codex~(code-davinci-002) as the LLM in our main experiments across ten datasets. 
Codex was provided for free when the paper was written, but the token consumption reduction is the same as the other LLMs, ensuring that the token costs in experiments are general.
We also test the batch prompting performance on other state-of-the-art LLMs, including GPT-3(text-davinci-003), GPT-3.5~(gpt-3.5-turbo), and GPT-4~(gpt-4). For GPT-4, we test the first $100$ samples for each dataset, considering the budget. 
The decoding temperature is set as $0$.
For each dataset, we manually select $12$-shot samples from the training set as in-context exemplars, with Chain-of-Thought~\cite[CoT]{cot} reasoning steps in the answers~(in Section~\ref{subsec:prompting methods}, other reasoning methods beyond CoT are discussed).
We choose $12$ exemplars because $12$ is the least common multiple of $2,3,4,6$, and thus it is easy to analyze the effects of grouping them into batches of $2,3,4,6$ samples in our ablation studies.
More experimental details and full results are listed in Appendix~\ref{appendix:full results}.

% ------------------------------------------------------------
\subsection{Main Results}

\begin{table}[t]
\small
\centering
\scalebox{0.95}{
    \begin{tabular}{l l c c}
    \toprule[1.2pt]
    \textbf{Task} & \textbf{Dataset}  & \textbf{Standard} & \textbf{Batch} \\
    \hline
    \textbf{Commonsense} & CSQA & $77.2$ & $\textbf{77.4}(+0.2)$ \\
    ~ & StrategyQA & $\textbf{73.3}$ & $71.0(-2.3)$ \\
    \midrule
    \textbf{Arithmetic} & GSM8K & $55.7$ & $\textbf{58.7}(+3.0)$ \\
    ~ & SVAMP & $\textbf{83.7}$ & $81.3(-2.4)$ \\
    ~ & AQuA & $\textbf{46.1}$ & $42.1(-4.0)$ \\
    ~ & AddSub & $\textbf{86.6}$ & $84.8(-1.8)$ \\
    ~ & MultiArith & $97.5$ & $\textbf{98.7}(+1.2)$ \\
    \midrule
    \textbf{NLI/NLU} & RTE & $\textbf{76.9}$ & $74.7(-2.2)$ \\
    ~ & MNLI & $65.3$ & $\textbf{65.7}(+0.4)$ \\
    ~ & SST-5 & $\textbf{51.3}$ & $49.7(-1.6)$ \\
    \bottomrule[1.2pt]
    \end{tabular} 
}
\caption{Accuracy of standard and batch prompting on ten datasets. Batch prompting shows comparable or even better performance. 
}
\label{tab:main performance}
\vspace{-0.3cm}
\end{table}

Figure~\ref{fig:main cost} compares the token and time costs of standard and batch prompting. 
As shown, batch prompting substantially~(up to $5\times$ with $6$ samples in each batch) reduces both the token and time costs of standard prompting with Codex. 
Further, the decrease of costs scales almost inverse linearly with the number of samples in each batch, verifying our analysis in Sections~\ref{subsec:token cost} and \ref{subsec:time cost}.
Note the time costs include the API call overhead and rate limit blocks, which exist in the commonly-used OpenAI and other LLM services.
For LLM services where these are not bottlenecks of time, the decoding time increase from larger $b$ should not be overlooked as discussed in Section~\ref{subsec:time cost}.
As the LLM infrastructure can change anytime, the token efficiency improvement is easier to compare than time; the token reduction in Figure~\ref{fig:main cost} should hold for any LLM over time.

Table~\ref{tab:main performance} shows that batch prompting~(with the best $b$, \textit{i.e.}, the number of samples in each batch) performs comparably or even better than standard prompting over all ten datasets.
We thus recommend that LLM users consider applying batch prompting to save money and time while maintaining good performance in realistic applications.

\subsection{Results across More LLMs}
\label{subsec:language models}

\begin{table}[t]
\centering
\small
\scalebox{0.865}{
    \begin{tabular}{l p{0.8cm}<{\centering} p{0.8cm}<{\centering} p{0.8cm}<{\centering} p{0.8cm}<{\centering} p{0.8cm}<{\centering} p{0.8cm}<{\centering}}
    \toprule[1.2pt]
    \multirow{2}{*}{\textbf{Dataset}} & \multicolumn{2}{c}{\textbf{GPT-3}}  & \multicolumn{2}{c}{\textbf{GPT-3.5}} & \multicolumn{2}{c}{\textbf{GPT-4}}\\
        ~ & {Standard} & {Batch} & {Standard} & {Batch} & {Standard} & {Batch} \\
    \hline
    CSQA & $78.3$ & $75.8$ & $72.9$ & $75.4$ & $84.0$ & $86.0$ \\
    GSM8K & $58.0$ & $55.0$ & $71.0$ & $76.7$ & $96.0$ & $93.0$ \\
    SVAMP & $86.7$ & $85.8$ & $84.7$ & $81.3$ & $98.0$ & $95.0$ \\
    AddSub & $99.2$ & $98.3$ & $89.3$ & $92.0$ & $99.0$ & $99.0$ \\
    RTE & $88.3$ & $88.3$ & $77.6$ & $81.6$ & $92.0$ & $90.0$ \\
    \bottomrule[1.2pt]
    \end{tabular} 
}
\caption{Accuracy of different LLMs with standard prompting and batch prompting using CoT prompts. Language models are GPT-3~(text-davinci-003), GPT-3.5~(gpt-3.5-turbo), and GPT-4~(gpt-4). Batch prompting can be applied well on different LLMs with good performance.
}
\label{tab:language models}
\vspace{-0.3cm}
\end{table}

We experiment batch prompting with some other state-of-the-art LLMs, including GPT-3, GPT-3.5~(ChatGPT) and GPT-4.

Table~\ref{tab:language models} shows performance from these LLMs.
All tested LLMs demonstrate capabilities similar to Codex: batch prompting retains downstream performance across datasets.
Actually, batch prompting Chat-based models tend to gain performance improvements.
We deduce the reason is that GPT-3.5 and GPT-4 accept a specific role of \textit{system message} as instruction, which makes them better follow batch prompting instructions to input and output in batches.
As discussed in Section~\ref{sec:approach}, the token efficiency of batch prompting should hold for different LLMs, though the decrease in time may vary depending on the LLM inference implementation.

% ----------------------------------------------------------------------------------------
\section{Analysis}
\label{sec:analysis}
In this section, we assess factors influencing batch prompting performance and the tradeoff between costs and performance. We also demonstrate that batch prompting can be applied to various LLM prompting methods, such as end-to-end and code generation.

% ------------------------------------------------------------
\subsection{Number of Batch Samples}

\begin{figure}[t]
    \begin{center}
    \includegraphics[width=2.9in]{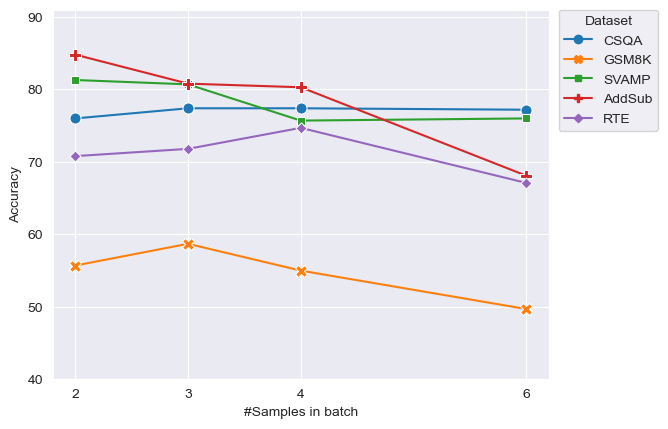}
    \end{center}
\caption{Accuracy over varying numbers of batch samples $b$ on five datasets using batch prompting. The performance decreases with larger $b$.}
\label{fig:number of samples in batch}
% \vspace{-0.3cm}
\end{figure}

Figure~\ref{fig:number of samples in batch} illustrates the impact of the number of samples per batch, $b$, on batch prompting performance. Performance typically decreases as $b$ increases, with a significant drop at $b=6$ across four out of five datasets. However, the optimal performance isn't always at $b=2$. Selecting $b=3$ or $b=4$ often yields good performance while conserving more tokens and time. The time/token cost reductions diminish as $b$ grows, suggesting $b<6$ (given 12 in-context examples in experiments) as a good balance between costs and performance.

% ------------------------------------------------------------
\subsection{Selection of Batch Samples}

\begin{table}[t]
\centering
\small
\scalebox{1.1}{
    \begin{tabular}{l c c c}
    \toprule[1.2pt]
    \textbf{Dataset} & \textbf{Random}  & \textbf{Similar} & \textbf{Diverse}\\
    \hline
    CSQA & $77.4$ & $77.4$ & $78.2$ \\
    GSM8K & $58.7$ & $57.7$ & $55.7$ \\
    SVAMP & $81.3$ & $81.3$ & $80.7$ \\
    AddSub & $84.8$ & $83.2$ & $84.1$ \\
    RTE & $74.7$ & $70.4$ & $70.8$ \\
    \bottomrule[1.2pt]
    \end{tabular} 
}
\caption{Accuracy from various batching methods on five representative datasets. Similarity or diversity-based methods do not achieve performance gains.
}
\label{tab:selection of samples in batch}
\vspace{-0.3cm}
\end{table}

Here we examine whether the selection of samples, \textit{i.e.} how samples are grouped into batches, will affect the performance of batch prompting.
We study two widely-adopted sample selection methods in in-context learning when grouping the test samples: grouping more similar~\cite{rubin2021learning,liu2022makes} and more diverse~\cite{su2022selective,Agrawal2022IncontextES} samples into batches.
Specifically, given $N$ test samples, to group similar ones, we use \textit{k-means clustering} and post-process each cluster into equal size $b$ by moving redundant samples to their closest groups with size $<b$. 
To group diverse ones, we apply the \textit{vote-k} method~\cite{su2022selective} to iteratively select diverse and representative groups of samples.

As listed in Table~\ref{tab:selection of samples in batch}, both similarity and diversity-based selections do not show improvements over random grouping.
We suspect that the reason may be that both methods assume in-batch samples can benefit from previous similar or diverse samples, \textit{i.e.}, samples in the front of the batch.
However, these earlier samples without ground truth outputs may bring error propagation to the rest of the in-batch samples.
%causing a performance drop.
Developing effective strategies for selecting samples for batch prompting could be a promising area for future research to further enhance the performance of batch prompting.

% ------------------------------------------------------------
\subsection{Complexity of Tasks}

\begin{figure}[t]
    \begin{center}
    \includegraphics[width=2.7in]{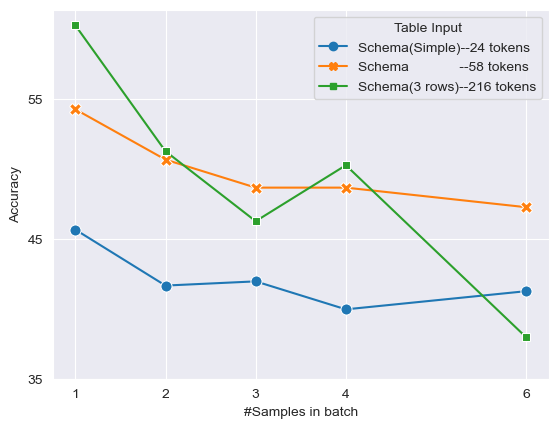}
    \end{center}
\caption{Accuracy on WikiTQ of various table input strategies and $b$~(the number of samples in each batch). This studies how the input length affects batch prompting performance.
$b\!=\!1$ means standard prompting.
Average input tokens per table are $24$, $58$, and $216$ tokens.
As the number of batch samples increases, batch prompting suffers in downstream performance. 
}
\label{fig:complexity of tasks}
\end{figure}

In Table~\ref{tab:main performance}, the steepest drop~(from $46.1$ to $42.1$) occurs on AQuA dataset: an arithmetic reasoning task in a multi-choice QA format.
One possible interpretation is that AQuA is more difficult than other datasets with the lowest absolute accuracy $46.1\%$, and thus LLMs are more likely to be disturbed when input contexts are grouped together.

We further study another task aspect that may affect performance: batch prompting tends to degrade performance more significantly with longer input contexts. 
% Since an AQuA example inputs both questions and answer choices, its input tokens are longer than simple QA.
We validate our assumption with WikiTQ~\cite{pasupat2015compositional},  a challenging Table QA dataset.
Tables contain longer input tokens for their multiple rows and columns.
We experiment with increasing table input lengths: a simplified table schema~(\textit{i.e.}, column names without column types; avg. $24$ tokens/table), a table schema~(avg. $58$ tokens/table), and a table schema with three table rows~(avg. $216$ tokens/table).
% We follow Binder~\cite{cheng2022binding} to generate Binder-SQL programs to solve the questions.

As shown in Figure~\ref{fig:complexity of tasks}, in standard prompting ($b\!=\!1$), inputting table schemas with three rows dominates QA performance.
However, it also sees the steepest performance drop when $b$ increases using batch prompting.
The shorter the input contexts, the steadier the performance with batch prompting.
This suggests that long task inputs are more likely to lead to confusion and performance drops when batch prompting is applied.

% ------------------------------------------------------------
\subsection{Reasoning Methods}
\label{subsec:prompting methods}

\begin{table}[t]
\centering
\small
\scalebox{1}{
    \begin{tabular}{l p{0.8cm}<{\centering} p{0.8cm}<{\centering}  p{0.8cm}<{\centering} p{0.8cm}<{\centering}}
    \toprule[1.2pt]
    \multirow{2}{*}{\textbf{Dataset}} & \multicolumn{2}{p{2cm}<{\centering}}{\textbf{End-to-end}}  & \multicolumn{2}{p{2cm}<{\centering}}{\textbf{Program}}\\
        ~ & {Standard} & {Batch} & {Standard} & {Batch} \\
    \hline
    CSQA & $81.5$ & $80.4$  & - & -\\
    GSM8K & $21.3$ & $17.3$ & $72.7$ & $73.0$\\
    SVAMP & $70.7$ & $68.3$ & $86.0$ & $86.3$ \\
    RTE & $85.2$ & $83.4$  & - & -\\
    WikiTQ & - & - & $54.3$ & $50.7$\\
    \bottomrule[1.2pt]
    \end{tabular} 
}
\caption{Accuracy of different reasoning methods with standard and batch prompting. Batch prompting can be applied well showing similar or better performance.
}
\label{tab:reasoning methods}
\vspace{-0.4cm}
\end{table}

In our main experiments~(Section \ref{sec:experiments}), we used the Chain-of-Thought~(CoT) for all ten datasets.
Here we examine whether batch prompting is suitable for other common LLM reasoning methods.
We experiment with two more reasoning methods: end-to-end (\textit{i.e.}, directly prompt the LLM to output the answers without intermediate steps) and program-based, (\textit{i.e.}, prompt the LLM to generate programs to answer the question).
For the program-based methods, we adopt Binder~\cite{cheng2022binding} on WikiTQ and Program-of-Thought~\cite[PoT]{chen2022program} on GSM8K and SVAMP.

As seen in Table~\ref{tab:reasoning methods}, both end-to-end and program-based methods can benefit from the efficiency of batch prompting while maintaining similar or even better performance on the task.
This indicates batch prompting is a drop-in replacement that can be combined with various reasoning methods under diverse scenarios.

% ------------------------------------------------------------

% ----------------------------------------------------------------------------------------
\section{Related Work}
\label{sec:related works}
\textbf{Improve In-Context Learning.}
The impressive capabilities of large language models~\cite[LLM]{gpt3,codex,palm} have sparked a surge of recent research aiming to enhance in-context learning~(ICL) performance.
Several works propose different reasoning methods to prompt LLMs~\cite{cot,zhou2022least,khot2022decomposed}, showing great improvements over directly prompting LLMs to output answers.
Other works~\cite{chen2022program,gao2022pal,cheng2022binding} generate programs to solve reasoning tasks. 
% Carefully designing prompting templates~\cite{jiang2020can,bach2022promptsource,arora2022ask} also benefit ICL.
Another line of work~\cite{liu2022makes,su2022selective,Agrawal2022IncontextES} focuses on selecting better in-context exemplars.
This work adds a new dimension to ICL for large-scale real-world applications: batch prompting to save budget and time while achieving good or even better performance.
%We believe that critical issue for LLM end-users.

\vspace{0.2cm}
\noindent \textbf{Efficient Language Generation.}
Much recent work proposed methods for efficient language generation, including machine translation \cite{Kasai2020ParallelMT,deepshallow,kasai2021t2r} and language modeling \cite{katharopoulos-et-al-2020,peng2021rfa,peng2021abc}, and model cascading~\cite{varshney2022model}.
Many of them introduce alternative architectures to the standard transformer to achieve such efficiency gains, which makes them hard to apply or deploy to real-world scenarios. Our method is a simple yet effective alternative to recent prompting methods, and thus it is applicable to any off-the-shelf language model APIs, such as OpenAI, Google, Anthropic, or any other available private LLM APIs, \emph{without} any additional training or customized model hosting.

% ----------------------------------------------------------------------------------------
% ----------------------------------------------------------------------------------------
\section{Limitation}
Batch prompting has proven to be an efficient method for time and token reduction. Nonetheless, there are several critical considerations to keep in mind when implementing it across various scenarios. \textbf{First, to optimize its benefits, the length of the input prompt tokens should be (significantly) greater than that of the output tokens.} Thus, it might not be suitable for ``heavy output" tasks like story generation. It is important to note that while our experiments are conducted with few-shot in-context learning, this method is also applicable to the instruction-following paradigm, either on its own or in combination, by simply substituting or adding the few-shot inputs with instructions. The only crucial factor is the length of the shared input tokens of inference samples. \textbf{Secondly, it is possible to observe performance declines.} Our experiments indicate that task complexity and lengthy input contexts can negatively impact performance. Although we have not identified a definitive guideline for predicting performance, we advise users to initiate testing with a smaller subset to gauge the effectiveness of batch prompting before implementing it on a larger scale.

% ----------------------------------------------------------------------------------------
% ----------------------------------------------------------------------------------------
\section{Conclusion}
\label{sec:conclusion}
We present batch prompting, a new way to prompt LLMs that performs inference on samples in a batched fashion.
With batch prompting, multiple samples can be handled in one API call so that the costs of tokens and time can be significantly reduced.
Extensive experiments on ten datasets across commonsense QA, arithmetics, and NLI/NLU 
% \jungo{mention these categories in the abstract too?}\zj{Added!} 
show that batch prompting can achieve better or similar performance compared to standard prompting, with much lower token and time costs.
%Different from previous prompting approaches that mostly focus on task performance, 
We hope batch prompting offers pragmatic value to efficient real-world LLM usage.

\clearpage

\bibliography{custom}

\begin{thebibliography}{38}
\expandafter\ifx\csname natexlab\endcsname\relax\def\natexlab#1{#1}\fi

\bibitem[{Agrawal et~al.(2022)Agrawal, Zhou, Lewis, Zettlemoyer, and Ghazvininejad}]{Agrawal2022IncontextES}
Sweta Agrawal, Chunting Zhou, Mike Lewis, Luke Zettlemoyer, and Marjan Ghazvininejad. 2022.
\newblock \href {https://arx iv.org/abs/2212.02437} {In-context examples selection for machine translation}.

\bibitem[{Arora et~al.(2022)Arora, Narayan, Chen, Orr, Guha, Bhatia, Chami, Sala, and R{\'e}}]{arora2022ask}
Simran Arora, Avanika Narayan, Mayee~F Chen, Laurel~J Orr, Neel Guha, Kush Bhatia, Ines Chami, Frederic Sala, and Christopher R{\'e}. 2022.
\newblock \href {https://arxiv.org/abs/2210.02441} {Ask me anything: A simple strategy for prompting language models}.

\bibitem[{Bach et~al.(2022)Bach, Sanh, Yong, Webson, Raffel, Nayak, Sharma, Kim, Bari, F{\'e}vry et~al.}]{bach2022promptsource}
Stephen Bach, Victor Sanh, Zheng~Xin Yong, Albert Webson, Colin Raffel, Nihal~V Nayak, Abheesht Sharma, Taewoon Kim, M~Saiful Bari, Thibault F{\'e}vry, et~al. 2022.
\newblock \href {https://arxiv.org/abs/2202.01279} {{PromptSource}: An integrated development environment and repository for natural language prompts}.
\newblock In \emph{Proc.\ of ACL}.

\bibitem[{Bentivogli et~al.(2009)Bentivogli, Clark, Dagan, and Giampiccolo}]{rte}
Luisa Bentivogli, Peter Clark, Ido Dagan, and Danilo Giampiccolo. 2009.
\newblock \href {https://link.springer.com/chapter/10.1007/11736790_9} {The fifth pascal recognizing textual entailment challenge}.
\newblock In \emph{Proc.\ of TAC}.

\bibitem[{Brown et~al.(2020)Brown, Mann, Ryder, Subbiah, Kaplan, Dhariwal, Neelakantan, Shyam, Sastry, Askell, Agarwal, Herbert-Voss, Krueger, Henighan, Child, Ramesh, Ziegler, Wu, Winter, Hesse, Chen, Sigler, Litwin, Gray, Chess, Clark, Berner, McCandlish, Radford, Sutskever, and Amodei}]{gpt3}
Tom Brown, Benjamin Mann, Nick Ryder, Melanie Subbiah, Jared~D Kaplan, Prafulla Dhariwal, Arvind Neelakantan, Pranav Shyam, Girish Sastry, Amanda Askell, Sandhini Agarwal, Ariel Herbert-Voss, Gretchen Krueger, Tom Henighan, Rewon Child, Aditya Ramesh, Daniel Ziegler, Jeffrey Wu, Clemens Winter, Chris Hesse, Mark Chen, Eric Sigler, Mateusz Litwin, Scott Gray, Benjamin Chess, Jack Clark, Christopher Berner, Sam McCandlish, Alec Radford, Ilya Sutskever, and Dario Amodei. 2020.
\newblock \href {https://proceedings.neurips.cc/paper/2020/file/1457c0d6bfcb4967418bfb8ac142f64a-Paper.pdf} {Language models are few-shot learners}.
\newblock In \emph{Proc.\ of NeurIPS}.

\bibitem[{Chen et~al.(2021)Chen, Tworek, Jun, Yuan, de~Oliveira~Pinto, Kaplan, Edwards, Burda, Joseph, Brockman, Ray, Puri, Krueger, Petrov, Khlaaf, Sastry, Mishkin, Chan, Gray, Ryder, Pavlov, Power, Kaiser, Bavarian, Winter, Tillet, Such, Cummings, Plappert, Chantzis, Barnes, Herbert-Voss, Guss, Nichol, Paino, Tezak, Tang, Babuschkin, Balaji, Jain, Saunders, Hesse, Carr, Leike, Achiam, Misra, Morikawa, Radford, Knight, Brundage, Murati, Mayer, Welinder, McGrew, Amodei, McCandlish, Sutskever, and Zaremba}]{codex}
Mark Chen, Jerry Tworek, Heewoo Jun, Qiming Yuan, Henrique~Ponde de~Oliveira~Pinto, Jared Kaplan, Harri Edwards, Yuri Burda, Nicholas Joseph, Greg Brockman, Alex Ray, Raul Puri, Gretchen Krueger, Michael Petrov, Heidy Khlaaf, Girish Sastry, Pamela Mishkin, Brooke Chan, Scott Gray, Nick Ryder, Mikhail Pavlov, Alethea Power, Lukasz Kaiser, Mohammad Bavarian, Clemens Winter, Philippe Tillet, Felipe~Petroski Such, Dave Cummings, Matthias Plappert, Fotios Chantzis, Elizabeth Barnes, Ariel Herbert-Voss, William~Hebgen Guss, Alex Nichol, Alex Paino, Nikolas Tezak, Jie Tang, Igor Babuschkin, Suchir Balaji, Shantanu Jain, William Saunders, Christopher Hesse, Andrew~N. Carr, Jan Leike, Josh Achiam, Vedant Misra, Evan Morikawa, Alec Radford, Matthew Knight, Miles Brundage, Mira Murati, Katie Mayer, Peter Welinder, Bob McGrew, Dario Amodei, Sam McCandlish, Ilya Sutskever, and Wojciech Zaremba. 2021.
\newblock \href {https://arxiv.org/abs/2107.03374} {Evaluating large language models trained on code}.

\bibitem[{Chen et~al.(2022)Chen, Ma, Wang, and Cohen}]{chen2022program}
Wenhu Chen, Xueguang Ma, Xinyi Wang, and William~W Cohen. 2022.
\newblock \href {https://arxiv.org/abs/2211.12588} {Program of thoughts prompting: Disentangling computation from reasoning for numerical reasoning tasks}.

\bibitem[{Cheng et~al.(2022)Cheng, Xie, Shi, Li, Nadkarni, Hu, Xiong, Radev, Ostendorf, Zettlemoyer, Smith, and Yu}]{cheng2022binding}
Zhoujun Cheng, Tianbao Xie, Peng Shi, Chengzu Li, R.K. Nadkarni, Yushi Hu, Caiming Xiong, Dragomir~R. Radev, Marilyn Ostendorf, Luke Zettlemoyer, Noah~A. Smith, and Tao Yu. 2022.
\newblock \href {https://arxiv.org/abs/2210.02875} {Binding language models in symbolic languages}.

\bibitem[{Chowdhery et~al.(2022)Chowdhery, Narang, Devlin, Bosma, Mishra, Roberts, Barham, Chung, Sutton, Gehrmann et~al.}]{palm}
Aakanksha Chowdhery, Sharan Narang, Jacob Devlin, Maarten Bosma, Gaurav Mishra, Adam Roberts, Paul Barham, Hyung~Won Chung, Charles Sutton, Sebastian Gehrmann, et~al. 2022.
\newblock \href {https://arxiv.org/abs/2204.02311} {{PaLM}: Scaling language modeling with pathways}.

\bibitem[{Cobbe et~al.(2021)Cobbe, Kosaraju, Bavarian, Chen, Jun, Kaiser, Plappert, Tworek, Hilton, Nakano et~al.}]{gsm8k}
Karl Cobbe, Vineet Kosaraju, Mohammad Bavarian, Mark Chen, Heewoo Jun, Lukasz Kaiser, Matthias Plappert, Jerry Tworek, Jacob Hilton, Reiichiro Nakano, et~al. 2021.
\newblock \href {https://arxiv.org/abs/2110.14168} {Training verifiers to solve math word problems}.

\bibitem[{Gao et~al.(2022)Gao, Madaan, Zhou, Alon, Liu, Yang, Callan, and Neubig}]{gao2022pal}
Luyu Gao, Aman Madaan, Shuyan Zhou, Uri Alon, Pengfei Liu, Yiming Yang, Jamie Callan, and Graham Neubig. 2022.
\newblock \href {https://arxiv.org/abs/2211.10435} {{PAL}: Program-aided language models}.

\bibitem[{Geva et~al.(2021)Geva, Khashabi, Segal, Khot, Roth, and Berant}]{strategyqa}
Mor Geva, Daniel Khashabi, Elad Segal, Tushar Khot, Dan Roth, and Jonathan Berant. 2021.
\newblock \href {https://arxiv.org/abs/2101.02235} {Did aristotle use a laptop? a question answering benchmark with implicit reasoning strategies}.
\newblock \emph{TACL}.

\bibitem[{Hosseini et~al.(2014)Hosseini, Hajishirzi, Etzioni, and Kushman}]{addsub}
Mohammad~Javad Hosseini, Hannaneh Hajishirzi, Oren Etzioni, and Nate Kushman. 2014.
\newblock \href {https://aclanthology.org/D14-1058/} {Learning to solve arithmetic word problems with verb categorization}.
\newblock In \emph{Proc.\ of EMNLP}.

\bibitem[{Jiang et~al.(2020)Jiang, Xu, Araki, and Neubig}]{jiang2020can}
Zhengbao Jiang, Frank~F Xu, Jun Araki, and Graham Neubig. 2020.
\newblock \href {https://arxiv.org/abs/1911.12543} {How can we know what language models know?}
\newblock \emph{TACL}.

\bibitem[{Kasai et~al.(2020)Kasai, Cross, Ghazvininejad, and Gu}]{Kasai2020ParallelMT}
Jungo Kasai, James Cross, Marjan Ghazvininejad, and Jiatao Gu. 2020.
\newblock \href {https://arxiv.org/abs/2001.05136} {Non-autoregressive machine translation with disentangled context transformer}.
\newblock In \emph{Proc.\ of ICML}.

\bibitem[{Kasai et~al.(2021{\natexlab{a}})Kasai, Pappas, Peng, Cross, and Smith}]{deepshallow}
Jungo Kasai, Nikolaos Pappas, Hao Peng, James Cross, and Noah~A. Smith. 2021{\natexlab{a}}.
\newblock \href {https://arxiv.org/abs/2006.10369} {Deep encoder, shallow decoder: Reevaluating non-autoregressive machine translation}.
\newblock In \emph{Proc.\ of ICLR}.

\bibitem[{Kasai et~al.(2021{\natexlab{b}})Kasai, Peng, Zhang, Yogatama, Ilharco, Pappas, Mao, Chen, and Smith}]{kasai2021t2r}
Jungo Kasai, Hao Peng, Yizhe Zhang, Dani Yogatama, Gabriel Ilharco, Nikolaos Pappas, Yi~Mao, Weizhu Chen, and Noah~A. Smith. 2021{\natexlab{b}}.
\newblock \href {https://arxiv.org/abs/2103.13076} {Finetuning pretrained transformers into {RNN}s}.
\newblock In \emph{Proc. of EMNLP}.

\bibitem[{Katharopoulos et~al.(2020)Katharopoulos, Vyas, Pappas, and Fleuret}]{katharopoulos-et-al-2020}
Angelos Katharopoulos, Apoorv Vyas, Nikolaos Pappas, and Fran\c{c}ois Fleuret. 2020.
\newblock \href {https://arxiv.org/abs/2006.16236} {Transformers are rnns: Fast autoregressive transformers with linear attention}.
\newblock In \emph{Proc.\ of ICML}.

\bibitem[{Khot et~al.(2022)Khot, Trivedi, Finlayson, Fu, Richardson, Clark, and Sabharwal}]{khot2022decomposed}
Tushar Khot, Harsh Trivedi, Matthew Finlayson, Yao Fu, Kyle Richardson, Peter Clark, and Ashish Sabharwal. 2022.
\newblock \href {https://arxiv.org/abs/2210.02406} {Decomposed prompting: A modular approach for solving complex tasks}.

\bibitem[{Ling et~al.(2017)Ling, Yogatama, Dyer, and Blunsom}]{aqua}
Wang Ling, Dani Yogatama, Chris Dyer, and Phil Blunsom. 2017.
\newblock \href {https://arxiv.org/abs/1705.04146} {Program induction by rationale generation: Learning to solve and explain algebraic word problems}.
\newblock In \emph{Proc.\ of ACL}.

\bibitem[{Liu et~al.(2022)Liu, Shen, Zhang, Dolan, Carin, and Chen}]{liu2022makes}
Jiachang Liu, Dinghan Shen, Yizhe Zhang, Bill Dolan, Lawrence Carin, and Weizhu Chen. 2022.
\newblock \href {https://arxiv.org/abs/2101.06804} {What makes good in-context examples for gpt-3?}

\bibitem[{Ouyang et~al.(2022)Ouyang, Wu, Jiang, Almeida, Wainwright, Mishkin, Zhang, Agarwal, Slama, Ray et~al.}]{ouyang2022training}
Long Ouyang, Jeffrey Wu, Xu~Jiang, Diogo Almeida, Carroll Wainwright, Pamela Mishkin, Chong Zhang, Sandhini Agarwal, Katarina Slama, Alex Ray, et~al. 2022.
\newblock Training language models to follow instructions with human feedback.
\newblock \emph{Advances in Neural Information Processing Systems}, 35:27730--27744.

\bibitem[{Pasupat and Liang(2015)}]{pasupat2015compositional}
Panupong Pasupat and Percy Liang. 2015.
\newblock \href {https://arxiv.org/abs/1508.00305} {Compositional semantic parsing on semi-structured tables}.
\newblock In \emph{Proc.\ of ACL}.

\bibitem[{Patel et~al.(2021)Patel, Bhattamishra, and Goyal}]{svamp}
Arkil Patel, Satwik Bhattamishra, and Navin Goyal. 2021.
\newblock \href {https://arxiv.org/abs/2103.07191} {Are {NLP} models really able to solve simple math word problems?}
\newblock In \emph{Proc.\ of NAACL}.

\bibitem[{Peng et~al.(2022)Peng, Kasai, Pappas, Yogatama, Wu, Kong, Schwartz, and Smith}]{peng2021abc}
Hao Peng, Jungo Kasai, Nikolaos Pappas, Dani Yogatama, Zhaofeng Wu, Lingpeng Kong, Roy Schwartz, and Noah~A. Smith. 2022.
\newblock \href {https://arxiv.org/abs/2110.02488} {{ABC}: Attention with bounded-memory control}.
\newblock In \emph{Proc. of ACL}.

\bibitem[{Peng et~al.(2021)Peng, Pappas, Yogatama, Schwartz, Smith, and Kong}]{peng2021rfa}
Hao Peng, Nikolaos Pappas, Dani Yogatama, Roy Schwartz, Noah Smith, and Lingpeng Kong. 2021.
\newblock \href {https://arxiv.org/abs/2103.02143} {Random feature attention}.
\newblock In \emph{Proc. of ICLR}.

\bibitem[{Roy and Roth(2015)}]{multiarith}
Subhro Roy and Dan Roth. 2015.
\newblock \href {https://arxiv.org/abs/1608.01413} {Solving general arithmetic word problems}.
\newblock In \emph{Proc.\ of EMNLP}.

\bibitem[{Rubin et~al.(2021)Rubin, Herzig, and Berant}]{rubin2021learning}
Ohad Rubin, Jonathan Herzig, and Jonathan Berant. 2021.
\newblock \href {https://arxiv.org/abs/2112.08633} {Learning to retrieve prompts for in-context learning}.

\bibitem[{Socher et~al.(2013)Socher, Perelygin, Wu, Chuang, Manning, Ng, and Potts}]{sst-5}
Richard Socher, Alex Perelygin, Jean Wu, Jason Chuang, Christopher~D Manning, Andrew~Y Ng, and Christopher Potts. 2013.
\newblock \href {https://nlp.stanford.edu/~socherr/EMNLP2013_RNTN.pdf} {Recursive deep models for semantic compositionality over a sentiment treebank}.
\newblock In \emph{Proc.\ of EMNLP}.

\bibitem[{Su et~al.(2022)Su, Kasai, Wu, Shi, Wang, Xin, Zhang, Ostendorf, Zettlemoyer, Smith, and Yu}]{su2022selective}
Hongjin Su, Jungo Kasai, Chen~Henry Wu, Weijia Shi, Tianlu Wang, Jiayi Xin, Rui Zhang, Mari Ostendorf, Luke Zettlemoyer, Noah~A Smith, and Tao Yu. 2022.
\newblock \href {https://arxiv.org/abs/2209.01975} {Selective annotation makes language models better few-shot learners}.

\bibitem[{Talmor et~al.(2019)Talmor, Herzig, Lourie, and Berant}]{commonsenseqa}
Alon Talmor, Jonathan Herzig, Nicholas Lourie, and Jonathan Berant. 2019.
\newblock \href {https://arxiv.org/abs/1811.00937} {{CommonsenseQA}: A question answering challenge targeting commonsense knowledge}.
\newblock In \emph{Proc.\ of NAACL}.

\bibitem[{Varshney and Baral(2022)}]{varshney2022model}
Neeraj Varshney and Chitta Baral. 2022.
\newblock Model cascading: Towards jointly improving efficiency and accuracy of nlp systems.
\newblock In \emph{Proceedings of the 2022 Conference on Empirical Methods in Natural Language Processing}, pages 11007--11021.

\bibitem[{Vaswani et~al.(2017)Vaswani, Shazeer, Parmar, Uszkoreit, Jones, Gomez, Kaiser, and Polosukhin}]{vaswani2017attention}
Ashish Vaswani, Noam Shazeer, Niki Parmar, Jakob Uszkoreit, Llion Jones, Aidan~N Gomez, {\L}ukasz Kaiser, and Illia Polosukhin. 2017.
\newblock \href {https://arxiv.org/abs/1706.03762} {Attention is all you need}.
\newblock In \emph{Proc.\ of NeurIPS}.

\bibitem[{Wang et~al.(2022)Wang, Wei, Schuurmans, Le, Chi, and Zhou}]{wang2022self}
Xuezhi Wang, Jason Wei, Dale Schuurmans, Quoc Le, Ed~Chi, and Denny Zhou. 2022.
\newblock \href {https://arxiv.org/abs/2203.11171} {Self-consistency improves chain of thought reasoning in language models}.

\bibitem[{Wei et~al.(2022)Wei, Wang, Schuurmans, Bosma, Chi, Le, and Zhou}]{cot}
Jason Wei, Xuezhi Wang, Dale Schuurmans, Maarten Bosma, Ed~Chi, Quoc Le, and Denny Zhou. 2022.
\newblock \href {https://arxiv.org/abs/2201.11903} {Chain of thought prompting elicits reasoning in large language models}.
\newblock In \emph{Proc.\ of NeurIPS}.

\bibitem[{Williams et~al.(2018)Williams, Nangia, and Bowman}]{mnli}
Adina Williams, Nikita Nangia, and Samuel Bowman. 2018.
\newblock \href {https://arxiv.org/abs/1704.05426} {A broad-coverage challenge corpus for sentence understanding through inference}.
\newblock In \emph{Proc.\ of NAACL}.

\bibitem[{Yao et~al.(2022)Yao, Zhao, Yu, Du, Shafran, Narasimhan, and Cao}]{yao2022react}
Shunyu Yao, Jeffrey Zhao, Dian Yu, Nan Du, Izhak Shafran, Karthik~R Narasimhan, and Yuan Cao. 2022.
\newblock React: Synergizing reasoning and acting in language models.
\newblock In \emph{The Eleventh International Conference on Learning Representations}.

\bibitem[{Zhou et~al.(2022)Zhou, Sch{\"a}rli, Hou, Wei, Scales, Wang, Schuurmans, Bousquet, Le, and Chi}]{zhou2022least}
Denny Zhou, Nathanael Sch{\"a}rli, Le~Hou, Jason Wei, Nathan Scales, Xuezhi Wang, Dale Schuurmans, Olivier Bousquet, Quoc Le, and Ed~Chi. 2022.
\newblock \href {https://arxiv.org/abs/2205.10625} {Least-to-most prompting enables complex reasoning in large language models}.

\end{thebibliography}
\bibliographystyle{acl_natbib}

\clearpage

\appendix
\section{Time Cost Analysis Regarding Transformer Architecture}
\label{appendix:time analysis}
In batch prompting, assume there are $K$ in-context exemplars~($C$ tokens per sample on average), $b$ samples in a batch to be inference.
Standard prompting is a special case where $b\!=\!1$.
Since most current LLMs~(\textit{e.g.},GPT-3, Codex, PaLM) are based on the Transformer decoder-only architecture, we focus on the time cost of the auto-regressive decoder.

The plain transformer time complexity for decoding one token is $O(n^2d)$, \textit{i.e.}, the time for encoding the embeddings of input tokens, where $n$ is the length of input tokens and $d$ is the dimension of embeddings. 
With the caching of previous tokens, the time complexity to decode each of the rest tokens is $O(nd)$.
We omit $d$ since it is a constant.
Thus, the time of one inference to decode $C\cdot b$ tokens:
\begin{equation}
    \begin{aligned}
        T_{encode} &= (CK)^2\\
        T_{decode} &= (CK+1) + \dots (CK+Cb) \\
        T &= T_{encode} + T_{decode}
    \end{aligned}
\end{equation}
where $T_{encode}$ is the time for encoding the input tokens in the decoder, and $T_{decode}$ is the time for decoding the rest tokens. $C$ can be seen as a constant. One inference time $T$ regarding $K$ and $b$ is:
\begin{equation}
    \begin{aligned}
        T &= C^2K^2 + Cb\cdot CK + \frac{Cb(Cb+1)}{2} \\
          &= C^2(K^2+bK+\frac{b^2}{2}) + \frac{Cb}{2}
    \end{aligned}
    \label{eq:time complexity}
\end{equation}
Thus, increasing $b$ in batch prompting will also increase the time cost of one inference.
The influence of $b$ also increases with its value and is relatively marginal when $b$ is small, especially when $b \ll K$, which is a common practice~($b\!=\!1$) in few-shot in-context learning. 

We can see a few examples by setting $K\!=\!12$~(as in experiments), $C\!=\!100$ with varying $b$ in Table~\ref{tab:time complexity} according to equation~\ref{eq:time complexity}.

\begin{table}[ht]
\centering
\scalebox{0.7}{
    \begin{tabular}{c c}
    \toprule[1.2pt]
    $\textbf{b}$ & \textbf{Time per inference} \\
    \hline
    \textbf{1} & $1565050$ \\
    \textbf{2} & $1700100$ \\
    \textbf{3} & $1845150$ \\
    \textbf{4} & $2000200$ \\
    \textbf{6} & $2340300$ \\
    \textbf{12} & $3600600$ \\
    \bottomrule[1.2pt]
    \end{tabular} 
}
\caption{Time(no unit) per inference with $K\!=\!12$, $C\!=\!100$ and various $b$.
}
\label{tab:time complexity}
\end{table}

Though the numbers are not accurate considering the constant coefficients of Big $O$ time complexity, we can learn the decoding time increase can not be overlooked as $b$ becomes large.
We do not emphasize this part in Section~\ref{subsec:time cost} because the overhead and rate limit blocking time of the OpenAI API make up the most proportion of time cost, and thus reducing the $N$ times of API calls to $N/b$ times almost inverse linearly reduce the time cost~(see Figure~\ref{fig:main cost}).

However, if the overhead and rate limits are no longer the bottlenecks, \textit{e.g.}, rate limits are strict for Codex~(code-davinci-002), GPT-3.5~(gpt-3.5-turbo) and GPT-4~(gpt-4) but not a big issue to GPT-3~(text-davinci-003), then the decoding time increase will be non-negligible.

% ----------------------------------------------------------------------------------------
% ----------------------------------------------------------------------------------------
\section{More Experimental Results}
\label{appendix:full results}
We list results for all experiments~(Tables~\ref{tab:full accuracy}-\ref{tab:pot}).
For the WikiTQ experiment with Binder, the LLM generation temperature is $0.4$ following its paper. For the other experiments, the temperature is $0$. For all experiments, top\_p $\;=\!1$, sampling\_n$\;=\!1$, logprobs$\;=\!1$, and stop\_tokens$\;=\!\backslash n \backslash n$. 
Five OpenAI keys are used as a polling pool on rotation to request the OpenAI API of Codex~(the rate limit errors still occur in the experiments and are counted into time cost since it is a practical issue).
If fewer OpenAI keys are used, there should be more rate limit errors because the request interval for one key will be shorter.

\begin{table*}[t]
\centering
\scalebox{1}{
    \begin{tabular}{l l c c c c c}
    \toprule[1.2pt]
    \multirow{2}{*}{\textbf{Task}} & \multirow{2}{*}{\textbf{Dataset}}  & \multirow{2}{*}{\textbf{Standard Prompting}} & \multicolumn{4}{c}{\textbf{Batch Prompting}}\\
    ~ & ~ & ~ & $b\!=\!2$ & $3$ & $4$ & $6$\\
    \hline
    \textbf{Commonsense} & CSQA & $77.2$ & $76.0$ & $77.4$ & $77.4$ & $77.2$\\
    ~ & StrategyQA & $73.3$ & $69.0$ & $67.7$ & $71.0$ & $67.7$ \\
    \midrule
    \textbf{Arithmetic} & GSM8K & $55.7$ & $55.7$ & $58.7$ & $55.0$ & $49.7$ \\
    ~ & SVAMP & $83.7$ & $81.3$ & $80.7$ & $75.7$ & $76.0$ \\
    ~ & AQuA & $46.1$ & $41.3$ & $42.1$ & $33.1$ & $37.4$ \\
    ~ & AddSub & $86.6$ & $84.8$ & $80.8$ & $80.3$ & $68.1$\\
    ~ & MultiArith & $97.5$ & $98.0$ & $98.7$ & $96.5$ & $96.3$ \\
    \midrule
    \textbf{NLI/NLU} & RTE & $76.9$ & $70.8$ & $71.8$ & $74.7$ & $67.1$ \\
    ~ & MNLI & $65.3$ & $65.7$ & $64.7$ & $65.3$ & $64.7$ \\
    ~ & SST-5 & $51.3$ & $48.0$ & $45.0$ & $49.7$ & $48.7$ \\
    \bottomrule[1.2pt]
    \end{tabular} 
}
\caption{Batch prompting accuracy with different $b$~(the number of samples in batch) compared with standard prompting on ten datasets. All use Codex~(code-davinci-002) as the LLM and Chain-of-Thought as the reasoning method.
}
\label{tab:full accuracy}
\end{table*}

\begin{table*}[t]
\centering
\scalebox{1}{
    \begin{tabular}{l l c c c c c}
    \toprule[1.2pt]
    \multirow{2}{*}{\textbf{Task}} & \multirow{2}{*}{\textbf{Dataset}}  & \multirow{2}{*}{\textbf{Standard Promting}} & \multicolumn{4}{c}{\textbf{Batch Prompting}}\\
    ~ & ~ & ~ & $b\!=\!2$ & $3$ & $4$ & $6$\\
    \hline
    \textbf{Commonsense} & CSQA & $7.37$ & $3.77$ & $2.57$ & $1.96$ & $1.40$\\
    ~ & StrategyQA & $7.62$ & $3.63$ & $2.85$ & $2.42$ & $1.99$ \\
    \midrule
    \textbf{Arithmetic} & GSM8K & $8.78$ & $4.55$ & $3.91$ & $3.75$ & $3.61$ \\
    ~ & SVAMP & $7.25$ & $3.69$ & $2.46$ & $2.50$ & $1.92$ \\
    ~ & AQuA & $7.02$ & $3.62$ & $2.60$ & $2.45$ & $1.77$ \\
    ~ & AddSub & $7.79$ & $4.32$ & $2.41$ & $1.58$ & $1.45$\\
    ~ & MultiArith & $6.80$ & $3.56$ & $2.51$ & $1.89$ & $1.38$ \\
    \midrule
    \textbf{NLI/NLU} & RTE & $6.50$ & $4.56$ & $2.73$ & $2.40$ & $1.29$ \\
    ~ & MNLI & $7.11$ & $3.78$ & $2.54$ & $2.22$ & $1.32$ \\
    ~ & SST-5 & $7.42$ & $3.23$ & $2.69$ & $2.22$ & $1.18$ \\
    \bottomrule[1.2pt]
    \end{tabular} 
}
\caption{Batch prompting time per sample with different $b$~(the number of samples in batch) compared with standard prompting on ten datasets. All use Codex~(code-davinci-002) as the LLM and Chain-of-Thought as the reasoning method.
}
\label{tab:full time}
\end{table*}

\begin{table*}[t]
\centering
\scalebox{1}{
    \begin{tabular}{l c c c c c}
    \toprule[1.2pt]
    \multirow{2}{*}{\textbf{Table Input}} & \multirow{2}{*}{\textbf{Standard Prompting}}  & \multicolumn{4}{c}{\textbf{Batch Prompting}}\\
    ~ & ~ & $b\!=\!2$ & $3$ & $4$ & $6$\\
    \hline
    Schema(Simple) & $45.7$ & $41.7$ & $42.0$ & $40.0$ & $41.3$ \\
    Schema & $54.3$ & $50.7$ & $48.7$ & $48.7$ & $47.3$ \\
    Schema($3$ table rows) & $60.3$ & $51.3$ & $46.3$ & $50.3$ & $38.0$\\
    \bottomrule[1.2pt]
    \end{tabular} 
}
\caption{Accuracy on WikiTQ of various table input strategies and $b$~(number of samples in batch) using Binder~\cite{cheng2022binding} to generate programs with Codex~(code-davinci-002).
}
\label{tab:binder}
\vspace{-0.3cm}
\end{table*}

\begin{table*}[t]
\centering
\scalebox{1}{
    \begin{tabular}{l c c c c c}
    \toprule[1.2pt]
    \multirow{2}{*}{\textbf{Dataset}} & \multirow{2}{*}{\textbf{Standard Prompting}}  & \multicolumn{4}{c}{\textbf{Batch Prompting}}\\
    ~ & ~ & $b\!=\!2$ & $3$ & $4$ & $6$\\
    \hline
    GSM8K & $72.7$ & $66.3$ & $70.7$ & $73.0$ & $51.5$ \\
    SVAMP & $86.0$ & $86.3$ & $83.0$ & $80.7$ & $84.3$ \\
    \bottomrule[1.2pt]
    \end{tabular} 
}
\caption{Accuracy on GSM8K and SVAMP with varying $b$~(number of samples in batch) using Program-of-Thought~\cite{chen2022program} to generate programs with Codex~(code-davinci-002).
}
\label{tab:pot}
\end{table*}

% ----------------------------------------------------------------------------------------
% ----------------------------------------------------------------------------------------
\section{Prompts}
\label{appendix:full prompts}
In the section, we list the prompt templates we use for each dataset~(Tables~\ref{tab:prompt csqa}-\ref{tab:prompt sst-5}). We follow CoT~\cite{cot} to build the prompts of CommonsenseQA, StrategyQA, GSM8K, SVAMP, AQuA, AddSub, MutliArith. 
We follow Binder~\cite{cheng2022binding} and Program-of-Thought~\cite{chen2022program} to build the prompts of WikiTQ, GSM8K (program), and SVAMP (program).
For RTE, MNLI, SST-5, we design the prompts ourselves using Chain-of-Thought.
For prompts with fewer than $12$ in-context exemplars, we manually add to $12$ samples using samples from the training set.
We show batch prompting prompts with $b=4$ as examples. 
For different $b$, we group the same $12$ samples according to $b$. 
When using ChatGPT in Section~\ref{subsec:language models}, the prompt format differs from Codex and GPT-3 because its conversational capability. See Table~\ref{tab:prompt chatgpt}.

\begin{table*}[t]
\centering
\scalebox{0.9}{
    \begin{tabular}{l}
    \toprule[1.2pt]
    \textbf{\underline{CommonsenseQA Prompt}}\\
    Q[1]: What do people use to absorb extra ink from a fountain pen? \\
    Answer Choices[1]: (a) shirt pocket (b) calligrapher’s hand (c) inkwell (d) desk drawer (e) blotter \\
    Q[2]: What home entertainment equipment requires cable? \\
    Answer Choices[2]: (a) radio shack (b) substation (c) television (d) cabinet \\
    Q[3]: The fox walked from the city into the forest, what was it looking for? \\
    Answer Choices[3]: (a) pretty flowers (b) hen house (c) natural habitat (d) storybook \\
    Q[4]: Sammy wanted to go to where the people were. Where might he go? \\
    Answer Choices[4]: (a) populated areas (b) race track (c) desert (d) apartment (e) roadblock \\
    A[1]: The answer must be an item that can absorb ink. Of the above choices, only blotters are used to \\ absorb ink. So the answer is (e). \\
    A[2]: The answer must require cable. Of the above choices, only television requires cable. So the answer\\ is (c). \\
    A[3]: The answer must be something in the forest. Of the above choices, only natural habitat is in the forest. \\ So the answer is (b). \\
    A[4]: The answer must be a place with a lot of people. Of the above choices, only populated areas have a\\ lot of people. So the answer is (a). \\
    \\
    Q[1]: Where do you put your grapes just before checking out? \\
    Answer Choices[1]: (a) mouth (b) grocery cart (c)supermarket (d) fruit basket (e) fruit market \\
    Q[2]: Google Maps and other highway and street GPS services have replaced what? \\
    Answer Choices[2]: (a) united states (b) mexico (c) countryside (d) atlas\\
    Q[3]: Before getting a divorce, what did the wife feel who was doing all the work?\\
    Answer Choices[3]: (a) harder (b) anguish (c) bitterness (d) tears (e) sadness\\
    Q[4]: James went to the tennis court that was located in his home what?\\
    Answer Choices[4]: (a) country club (b) park (c) michigan (d) sports (e) town\\
    A[1]: The answer should be the place where grocery items are placed before checking out. Of the above \\ choices, grocery cart makes the most sense for holding grocery items. So the answer is (b).\\
    A[2]: The answer must be something that used to do what Google Maps and GPS services do, which is to \\ give directions. Of the above choices, only atlases are used to give directions. So the answer is (d).\\
    A[3]: The answer should be the feeling of someone getting divorced who was doing all the work. Of the \\ above choices, the closest feeling is bitterness. So the answer is (c).\\
    A[4]: The answer must be a place where tennis courts are located. Of the above choices, only home town \\has tennis courts. So the answer is (e).\\
    \\
    Q[1]: What does you body do when you exercise? \\
    Answer Choices[1]: (a) need for food (b) thirst (c) work out (d) sweating (e) injury\\
    Q[2]: In order to see a story on the big screen what must you do?\\
    Answer Choices[2]: (a) go to movies (b) visualize (c) reading (d) open book (e) sketching a picture\\
    Q[3]: He followed the train tracks hoping to get home, he had gotten lost in the Yooperland where?\\
    Answer Choices[3]: (a) ghetto (b) michigan (c) new york (d) canada (e) train station\\
    Q[4]: What would you get if you want a painting but cannot afford the original?\\
    Answer Choices[4]: (a) reproduction (b) derivative (c) reproduction (d) simile (e) remake\\
    A[1]: The answer must be something that happens when you exercise. Of the above choices, only sweating \\happens when you exercise. So the answer is (d).\\
    A[2]: The answer must be something that you do to see a story on the big screen. Of the above choices, \\only going to movies makes sense. So the answer is (a).\\
    A[3]: The answer should be a place that relates to Yooperland. Of the above choices, only michigan is \\related to Yooperland. So the answer is (b).\\
    A[4]: The answer must be something that is similar to the original. Of the above choices, only \\reproduction is similar to the original. So the answer is (a).\\
    \bottomrule[1.2pt]
    \end{tabular} 
}
\caption{CommonsenseQA Prompt.
}
\label{tab:prompt csqa}
\end{table*}

\begin{table*}[t]
\centering
\scalebox{0.9}{
    \begin{tabular}{l}
    \toprule[1.2pt]
    \textbf{\underline{StrategyQA Prompt}}\\
    Q[1]: Do hamsters provide food for any animals? \\
    Q[2]: Could Brooke Shields succeed at University of Pennsylvania?\\
    Q[3]: Hydrogen’s atomic number squared exceeds number of Spice Girls?\\
    Q[4]: Is it common to see frost during some college commencements?\\
    A[1]: Hamsters are prey animals. Prey are food for predators. Thus, hamsters provide food for some \\animals. So the answer is yes.\\
    A[2]: Brooke Shields went to Princeton University. Princeton University is about as academically \\rigorous as the University of Pennsylvania. Thus, Brooke Shields could also succeed at the University of \\Pennsylvania. So the answer is yes.\\
    A[3]: Hydrogen has an atomic number of 1. 1 squared is 1. There are 5 Spice Girls. Thus, Hydrogen’s \\atomic number squared is less than 5. So the answer is no.\\
    A[4]: College commencement ceremonies can happen in December, May, and June. December is in the \\winter, so there can be frost. Thus, there could be frost at some commencements. So the answer is yes.\\
    \\
    Q[1]: Could a llama birth twice during War in Vietnam (1945-46)?\\
    Q[2]: Would a pear sink in water?\\
    Q[3]: Can an Arvanite Greek understand some of the Albanian Declaration of Independence?\\
    Q[4]: Can Burundi's communicate with citizens of New Brunswick?\\
    A[1]: The War in Vietnam was 6 months. The gestation period for a llama is 11 months, which is more than \\6 months. Thus, a llama could not give birth twice during the War in Vietnam. So the answer is no.\\
    A[2]: The density of a pear is about 0.6g/cm3, which is less than water. Objects less dense than water \\float. Thus, a pear would float. So the answer is no.\\
    A[3]: The Arvanite Greek's are a major Tosk speaking group of southern Albania. Thus, they can understand\\ some of the Albanian Declaration of Independence. So the answer is yes.\\
    A[4]: French is one of the official languages of Burundi. Thus, Burundi's can communicate with citizens of \\New Brunswick. So the answer is yes.\\
    \\
    Q[1]: Are quadrupeds represented on Chinese calendar?\\
    Q[2]: Can actress Dafne Keen win the Eurovision Song Contest finals in 2020?\\
    Q[3]: Would a student in eleventh grade be unable to run for president of the United States?\\
    Q[4]: Does the judo rank system reach the triple digits?\\
    A[1]: The Chinese calendar has a number of symbols including monkeys, goats, and tigers. Tigers have four\\ paws and balance themselves by walking on their toes. Thus, quadrupeds are represented on the Chinese\\ calendar. So the answer is yes.\\
    A[2]: Contestants must be at least 16 years of age to compete in the finals of Eurovision Song Contest.\\ Dafne Keen is 15 years old in 2020. Thus, Dafne Keen cannot win the Eurovision Song Contest finals in 2020.\\ So the answer is no.\\
    A[3]: Students in the eleventh grade are typically 16–17 years of age. To serve as president, one must be at\\ least 35 years old. Thus, a student in eleventh grade would be unable to run for president of the United States.\\ So the answer is yes.\\
    A[4]: A triple digit number would be equal to at least 100. The judo dan-rank system was capped at 10th \\dan after the death of judo's founder, Kanō Jigorō. Thus, the judo rank system does not reach the triple \\digits. So the answer is no.\\
    \bottomrule[1.2pt]
    \end{tabular} 
}
\caption{StrategyQA Prompt.
}
\label{tab:prompt strategyqa}
\end{table*}

\begin{table*}[t]
\centering
\scalebox{0.82}{
    \begin{tabular}{l}
    \toprule[1.2pt]
    \textbf{\underline{GSM8K, SVAMP, AddSub, MultiArith Prompt}}\\
    Q[1]: There are 15 trees in the grove. Grove workers will plant trees in the grove today. After they are done, \\there will be 21 trees. How many trees did the grove workers plant today?\\
    Q[2]: If there are 3 cars in the parking lot and 2 more cars arrive, how many cars are in the parking lot?\\
    Q[3]: Leah had 32 chocolates and her sister had 42. If they ate 35, how many pieces do they have left \\in total?\\
    Q[4]: Jason had 20 lollipops. He gave Denny some lollipops. Now Jason has 12 lollipops. How many lollipops\\ did Jason give to Denny?\\
    A[1]: There are 15 trees originally. Then there were 21 trees after some more were planted. So there must have\\ been 21 - 15 = 6. The answer is 6.\\
    A[2]: There are originally 3 cars. 2 more cars arrive. 3 + 2 = 5. The answer is 5.\\
    A[3]: Originally, Leah had 32 chocolates. Her sister had 42. So in total they had 32 + 42 = 74. After eating 35,\\ they had 74 - 35 = 39. The answer is 39.\\
    A[4]: Jason started with 20 lollipops. Then he had 12 after giving some to Denny. So he gave Denny 20 - 12 \\= 8. The answer is 8.\\
    \\
    Q[1]: Shawn has five toys. For Christmas, he got two toys each from his mom and dad. How many toys does he \\have now?\\
    Q[2]: There were nine computers in the server room. Five more computers were installed each day, from monday\\ to thursday. How many computers are now in the server room?\\
    Q[3]: Michael had 58 golf balls. On tuesday, he lost 23 golf balls. On wednesday, he lost 2 more. How many golf\\ balls did he have at the end of wednesday?\\
    Q[4]: Olivia has \$23. She bought five bagels for \$3 each. How much money does she have left?\\
    A[1]: Shawn started with 5 toys. If he got 2 toys each from his mom and dad, then that is 4 more toys. 5 + 4 = 9. \\The answer is 9.\\
    A[2]: There were originally 9 computers. For each of 4 days, 5 more computers were added. So 5 * 4 = 20 \\computers were added. 9 + 20 is 29. The answer is 29.\\
    A[3]: Michael started with 58 golf balls. After losing 23 on tuesday, he had 58 - 23 = 35. After losing 2 more, he had \\35 - 2 = 33 golf balls. The answer is 33.\\
    A[4]: Olivia had 23 dollars. 5 bagels for 3 dollars each will be 5 x 3 = 15 dollars. So she has 23 - 15 dollars left. \\23 - 15 is 8. The answer is 8.\\
    \\
    Q[1]: A garden produced 237 potatoes, 60 fewer cucumbers and twice as many peppers than the cucumbers. How \\many vegetables did the garden produce?\\
    Q[2]: John's cow weighs 400 pounds. It increased its weight to 1.5 times its starting weight. He is able to sell the cow\\ for \$3 per pound. How much more is it worth after gaining the weight?\\
    Q[3]: John writes 20 pages a day. How long will it take him to write 3 books that are 400 pages each?\\
    Q[4]: James has a rainwater collection barrel. For each inch of rain he collects 15 gallons. On Monday it rained 4 inches\\ and on Tuesday it rained 3 inches. He can sell water for \$1.2 per gallon. How much money did he make from selling\\ all the water?\\
    A[1]: The garden produced 237 - 60 = 177 cucumbers. The garden produced 177 * 2 = 354 peppers. The garden \\produced 237 + 177 + 354 = 768 vegetables. The answer is 768.\\
    A[2]: The cow initially weighs 400 * 1.5 = 600 pounds. So it gained 600 - 400 = 200 pounds. It is worth 200 * 3 = 600 \\dollars more. The answer is 600.\\
    A[3]: He wants to write 3 * 400 = 1200 pages. So it will take him 1200 / 20= 60 days The answer is 60.\\
    A[4]: It rained 3 + 4 = 7 inches So he collected 7 * 15 = 105 gallons So he makes 105 * 1.2 = 126 from selling the \\water. The answer is 126.\\
    \bottomrule[1.2pt]
    \end{tabular} 
}
\caption{GSM8K, SVAMP, AddSub, MultiArith Prompt.
}
\label{tab:prompt gsm8k}
\end{table*}

\begin{table*}[t]
\centering
\scalebox{0.85}{
    \begin{tabular}{l}
    \toprule[1.2pt]
    \textbf{\underline{AQuA Prompt}}\\
    Q[1]: John found that the average of 15 numbers is 40. If 10 is added to each number then the mean of the \\numbers is?
    Answer Choices[1]: (a) 50 (b) 45 (c) 65 (d) 78 (e) 64\\
    Q[2]: If a / b = 3/4 and 8a + 5b = 22,then find the value of a.\\
    Answer Choices[2]: (a) 1/2 (b) 3/2 (c) 5/2 (d) 4/2 (e) 7/2\\
    Q[3]: A person is traveling at 20 km/hr and reached his destiny in 2.5 hr then find the distance?\\
    Answer Choices[3]: (a) 53 km (b) 55 km (c) 52 km (d) 60 km (e) 50 km\\
    Q[4]: How many keystrokes are needed to type the numbers from 1 to 500?\\
    Answer Choices[4]: (a) 1156 (b) 1392 (c) 1480 (d) 1562 (e) 1788\\
    A[1]: If 10 is added to each number, then the mean of the numbers also increases by 10. So the new mean \\would be 50. The answer is (a).\\
    A[2]: If a / b = 3/4, then b = 4a / 3. So 8a + 5(4a / 3) = 22. This simplifies to 8a + 20a / 3 = 22, which means\\ 44a / 3 = 22. So a is equal to 3/2. The answer is (b).\\
    A[3]: The distance that the person traveled would have been 20 km/hr * 2.5 hrs = 50 km. The answer is (e).\\
    A[4]: There are 9 one-digit numbers from 1 to 9. There are 90 two-digit numbers from 10 to 99. There are \\401 three-digit numbers from 100 to 500. 9 + 90(2) + 401(3) = 1392. The answer is (b).\\
    \\
    Q[1]: A number X equals 80\% of the average of 5, 7, 14 and a number Y. If the average of X and Y is 26, the \\value of Y is?\\
    Answer Choices[1]: (a) 13 (b) 26 (c) 39 (d)36 (e) None of these\\
    Q[2]: A shopkeeper gave an additional 20 per cent concession on the reduced price after giving 30 per \\cent standard concession on an article. If Arun bought that article for 1,120, what was the original price?\\
    Answer Choices[2]: (a) 3,000 (b) 4,000 (c) 2,400 (d) 2,000 (e) None of these\\
    Q[3]: A and B invests Rs.3000 and Rs.7000 respectively in a business. If A doubles his capital after 6 months.\\ In what ratio should A and B divide that year's profit?\\
    Answer Choices[3]: (a) 9:6 (b) 9:8 (c) 9:14 (d) 9:9 (e) 9:5\\
    Q[4]: The angle between two hands at 3.45 is?\\
    Answer Choices[4]: (a) 110 degree (b) 115 degree (c) 112 1/2 degree (d) 117 degree (e) 157 1/2 degree\\
    A[1]: Average of 5, 7, 14 and Y = (5 + 7 + 14 + Y) / 4. Therefore, X = 80\% of (5 + 7 + 14 + y) / 4 = (80/100)\\ x (26 + Y)/4 => X = (26 + Y)/5, i.e., 5X - Y = 26. Also, (X + Y) / 2 = 26. Thus, (26 + Y) / 5 + Y = 52, then Y \\= 39. The answer is (c).\\
    A[2]: The total discount should be (1 - 0.3) * (1 - 0.2) = 0.56. Thus, the original price should be 1120 / 0.56 \\= 2000. The answer is (d).\\
    A[3]: The ratio should be (3 * 6 + 6 * 6): (7 * 12) = 54:84. It simplifies to 9:14. The answer is (c).\\
    A[4]: The hour hand is (45/60) * (360/12) = 22.5 degree from 3 o'clock. So the angle between the hour hand and\\ the minute hand is (9-3) * (360/12) - 22.5 = 157.5. The answer is (e).\\
    \\
    Q[1]: Find the sum of first 30 natural numbers.\\
    Answer Choices[1]: (a) 470 (b) 468 (c) 465 (d) 463 (e) 487\\
    Q[2]: What will come in place of the x in the following Number series? 46080, 3840, ?, 48, 8, 2, 1.\\
    Answer Choices[2]: (a) 1 (b) 384 (c) 5 (d) 7 (e) 9\\
    Q[3]: A password of a computer used two digits where they are from 0 and 9. What is the probability that the \\password solely consists of prime numbers and zero?\\
    Answer Choices[3]: (a) 1/32 (b) 1/16 (c) 1/8 (d) 2/5 (e) 1/4\\
    Q[4]: If $k^3$ is divisible by 120, what is the least possible value of integer k?\\
    Answer Choices[4]: (a) 12 (b) 30 (c) 60 (d) 90 (e) 120\\
    A[1]: The sum of first 30 natural numbers is 30 * (30 + 1) / 2 = 465. The answer is (c).\\
    A[2]: The ratio of the numbers is 10:8:6:4:2:1. So the next number should be 384. The answer is (b).\\
    A[3]: 0, 2, 3, 5, 7 are five prime digits(including zero). So there are 5 * 5 = 25 two-digit numbers with \\only prime numbers and zero. The probability is 25/100 = 1/4. The answer is (e).\\
    A[4]: 120 can be factored as 2 * 2 * 2 * 3 * 5. So the least k be 2 * 3 * 5 = 30. The answer is (b).\\
    \bottomrule[1.2pt]
    \end{tabular} 
}
\caption{AQuA Prompt.
}
\label{tab:prompt aqua}
\end{table*}

\begin{table*}[t]
\centering
\scalebox{0.85}{
    \begin{tabular}{l}
    \toprule[1.2pt]
    \textbf{\underline{RTE Prompt}}\\
    Premise[1]: No Weapons of Mass Destruction Found in Iraq Yet.\\
    Hypothesis[1]: Weapons of Mass Destruction Found in Iraq.\\
    Premise[2]: A place of sorrow, after Pope John Paul II died, became a place of celebration, as Roman\\ Catholic faithful gathered in downtown Chicago to mark the installation of new Pope Benedict XVI.\\
    Hypothesis[2]: Pope Benedict XVI is the new leader of the Roman Catholic Church.\\
    Premise[3]: Libya's case against Britain and the US concerns the dispute over their demand \\for extradition of Libyans charged with blowing up a Pan Am jet over Lockerbie in 1988.\\
    Hypothesis[3]: One case involved the extradition of Libyan suspects in the Pan Am Lockerbie bombing.\\
    Premise[4]: Argentina sought help from Britain on its privatization program and encouraged British \\investment.\\
    Hypothesis[4]: Argentina sought UK expertise on privatization and agriculture.\\
    Answer[1]: No Weapons of Mass Destruction Found, which contradicts the hypothesis. So the\\ answer is False.\\
    Answer[2]: As Roman Catholic faithful gathered in downtown Chicago to mark the installation of new \\Pope Benedict XVI. So the answer is True.\\
    Answer[3]: Libya's case suspects in the Pan Am Lockerbie bombing. So the answer is True.\\
    Answer[4]: Argentina sought help from Britain on its privatization program, not agriculture, which \\contradicts the hypothesis. So the answer is False.\\
    \\
    Premise[1]: Startling new research into mobile phones claims they may reduce a man's sperm count by \\up to 30\%.\\
    Hypothesis[1]: Male fertility may be affected by use of a mobile phones.\\
    Premise[2]: It rewrites the rules of global trade, established by the General Agreement on Tariffs and \\Trade, or GATT, in 1947, and modified in multiple rounds of negotiations since then.\\
    Hypothesis[2]: GATT was formed in 1947.\\
    Premise[3]: The cost of the consumer of the United States fell in June.\\
    Hypothesis[3]: U.S. consumer spending dived in June.\\
    Premise[4]: Israeli Prime Minister Ariel Sharon has said that Mahmoud Abbas is a man that Israel can do\\ business with.
    Hypothesis[4]: Palestinian leader, Mahmoud Abbas, may be someone Israel can talk with.\\
    Answer[1]: New research claims mobile phones reduce a man's sperm count, i.e., affects male fertility. So \\the answer is True.\\
    Answer[2]: GATT is rewritten in 1947, not formed in 1947, which contradicts the hypothesis. So the answer\\ is False.\\
    Answer[3]: The consumer cost fell in June, not the spending, which contradicts the hypothesis. So the \\answer is False.\\
    Answer[4]: Mahmoud Abbas is a man that Israel can do business with, i.e., he may be someone Israel\\ can talk with. So the answer is True.\\
    \\
    Premise[1]: In October, however, amid rising tensions between the government and opposition groups, \\a car bomb seriously injured an opposition politician and killed his driver, in Beirut.\\
    Hypothesis[1]: A member of the opposition was injured in a car bomb attack in Beirut.\\
    Premise[2]: Ruth's 1927 single season record of 60 home runs stood unsurpassed until Roger Maris hit 61 in 1961.\\
    Hypothesis[2]: Babe Ruth hit 60 home runs in his lifetime.\\
    Premise[3]: The German technology was employed to build Shanghai's existing maglev line, the first \\in the world to be used commercially.\\
    Hypothesis[3]: Maglev is commercially used.\\
    Premise[4]: Twelve of Jupiter's moons are relatively small and seem to have been more likely captured \\than to have been formed in orbit around Jupiter.\\
    Hypothesis[4]: Jupiter has Twelve moons.\\
    Answer[1]: A car bomb seriously injured an opposition politician in Beirut. So the answer the True.\\
    Answer[2]: Babe Ruth hit 60 home runs in a single season, not his lifetime, which contradicts the hypothesis.\\ So the answer is False.\\
    Answer[3]: The German technology was employed to build Shanghai's existing maglev line, i.e., Maglev \\is commercially used. So the answer is True.\\
    Answer[4]: Twelve of Jupiter's moons are relatively small, not Jupiter has Twelve moons, which contradicts\\ the hypothesis. So the answer is False.\\
    \bottomrule[1.2pt]
    \end{tabular} 
}
\caption{RTE Prompt.
}
\label{tab:prompt rte}
\end{table*}

\begin{table*}[t]
\centering
\scalebox{0.9}{
    \begin{tabular}{l}
    \toprule[1.2pt]
    \textbf{\underline{MNLI Prompt}}\\
    Premise[1]: Conceptually cream skimming has two basic dimensions - product and geography.\\
    Hypothesis[1]: Product and geography are what make cream skimming work.\\
    Premise[2]: One of our number will carry out your instructions minutely.\\
    Hypothesis[2]: A member of my team will execute your orders with immense precision.\\
    Premise[3]: Analyzing Postal Service accounts for depreciation, fuel, and maintenance for \\city delivery carriers, we have estimated the average city delivery vehicle cost per route.\\
    Hypotheis[3]: Driving cost estimates can be averaged with sufficient data.\\
    Premise[4]: Consider the United States Postal Service.\\
    Hypothesis[4]: Forget the United States Postal Service.\\
    Answer[1]: The answer is Neutral.\\
    Answer[2]: The answer is True.\\
    Answer[3]: The answer is Neutral.\\
    Answer[4]: The answer is False.\\
    \\
    Premise[1]: Take a remarkable statistic that Shesol cites but lets pass relatively unexamined.\\
    Hypothesis[1]: They had data that was very relevant but under used.\\
    Premise[2]: The man on the ground thinks for a moment and yells back, You must work in management.\\
    Hypothesis[2]: There was no one on the ground, man or woman.\\
    Premise[3]: Hello, Ben.\\
    Hypothesis[3]: I ignored Ben.\\
    Premise[4]: How can you prove it?\\
    Hypothesis[4]: Can you tell me how to prove it?\\
    Answer[1]: The answer is True.\\
    Answer[2]: The answer is False.\\
    Answer[3]: The answer is False.\\
    Answer[4]: The answer is True.\\
    \\
    Premise[1]: In the midst of this amazing amalgam of cultures is a passion for continuity.\\
    Hypothesis[1]: A passion for continuity is not the most important of these cultures.\\
    Premise[2]: Poirot, I exclaimed, with relief, and seizing him by both hands, I dragged him into the room.\\
    Hypothesis[2]: Poirot was now back and I was sorry that he would take over what I now considered \\my own investigation.\\
    Premise[3]: There's a uh a couple called um oh i'm going to forgot his name now uh Dirkson.\\
    Hypothesis[3]: I can't remember their name.\\
    Premise[4]: It's not that the questions they asked weren't interesting or legitimate (though most did fall \\under the category of already asked and answered).\\
    Hypothesis[4]: All of the questions were interesting according to a focus group consulted on the subject.\\
    Answer[1]: The answer is Neutral.\\
    Answer[2]: The answer is False.\\
    Answer[3]: The answer is True.\\
    Answer[4]: The answer is Neutral.\\
    \bottomrule[1.2pt]
    \end{tabular} 
}
\caption{MNLI Prompt.
}
\label{tab:prompt mnli}
\end{table*}

\begin{table*}[t]
\centering
\scalebox{0.9}{
    \begin{tabular}{l}
    \toprule[1.2pt]
    \textbf{\underline{SST-5 Prompt}}\\    
    Q[1]: a stirring , funny and finally transporting re-imagining of beauty and the beast and 1930s \\horror films.\\
    Q[2]: they presume their audience wo n't sit still for a sociology lesson, however entertainingly\\ presented, so they trot out the conventional science-fiction elements of bug-eyed monsters and \\futuristic women in skimpy clothes.\\
    Q[3]: um , no..\\
    Q[4]: jonathan parker's bartleby should have been the be-all-end-all of the modern-office anomie films.\\
    A[1]: The tone is very positive.\\
    A[2]: The tone is negative.\\
    A[3]: The tone is neutral.\\
    A[4]: The tone is positive.\\
    \\
    Q[1]: lacks the inspiration of the original and has a bloated plot that stretches the running time \\about 10 minutes past a child's interest and an adult's patience.\\
    Q[2]: the santa clause 2 proves itself a more streamlined and thought out encounter than the original\\ could ever have hoped to be.\\
    Q[3]: you might say tykwer has done all that heaven allows, if you wanted to make as anti-kieslowski\\ a pun as possible.\\
    Q[4]: otto-sallies has a real filmmaker's eye.\\
    A[1]: The tone is very negative.\\
    A[2]: The tone is positive.\\
    A[3]: The tone is neutral.\\
    A[4]: The tone is positive.\\
    \\
    Q[1]: with a confrontational stance, todd solondz takes aim on political correctness and suburban\\ families.\\
    Q[2]: verall , cletis tout is a winning comedy that excites the imagination and tickles the funny bone.\\
    Q[3]: with its parade of almost perpetually wasted characters ... margarita feels like a hazy high \\that takes too long to shake.\\
    Q[4]: an ugly-duckling tale so hideously and clumsily told it feels accidental.\\
    A[1]: The tone is neutral.\\
    A[2]: The tone is very positive.\\
    A[3]: The tone is negative.\\
    A[4]: The tone is very negative.\\
    \bottomrule[1.2pt]
    \end{tabular} 
}
\caption{SST-5 Prompt.
}
\label{tab:prompt sst-5}
\end{table*}

\begin{table*}[t]
\centering
\scalebox{0.72}{
    \begin{tabular}{l}
    \toprule[1.2pt]
    \textbf{\underline{Chat Prompt for GSM8K}}\\    
    \textcolor{cyan}{System} \\
    The answers should respond to each question the user inputs. One line one answer starting with A[{idx}],\\ no other line break.\\
    \textcolor{violet}{User}\\
    Q[1]: There are 15 trees in the grove. Grove workers will plant trees in the grove today. After they are done, \\there will be 21 trees. How many trees did the grove workers plant today?\\
    Q[2]: If there are 3 cars in the parking lot and 2 more cars arrive, how many cars are in the parking lot?\\
    Q[3]: Leah had 32 chocolates and her sister had 42. If they ate 35, how many pieces do they have left \\in total?\\
    Q[4]: Jason had 20 lollipops. He gave Denny some lollipops. Now Jason has 12 lollipops. How many lollipops\\ did Jason give to Denny?\\
    \textcolor{brown}{Assistant}\\
    A[1]: There are 15 trees originally. Then there were 21 trees after some more were planted. So there must have\\ been 21 - 15 = 6. The answer is 6.\\
    A[2]: There are originally 3 cars. 2 more cars arrive. 3 + 2 = 5. The answer is 5.\\
    A[3]: Originally, Leah had 32 chocolates. Her sister had 42. So in total they had 32 + 42 = 74. After eating 35,\\ they had 74 - 35 = 39. The answer is 39.\\
    A[4]: Jason started with 20 lollipops. Then he had 12 after giving some to Denny. So he gave Denny 20 - 12 \\= 8. The answer is 8.\\
    \textcolor{violet}{User}\\
    Q[1]: Shawn has five toys. For Christmas, he got two toys each from his mom and dad. How many toys does he \\have now?\\
    Q[2]: There were nine computers in the server room. Five more computers were installed each day, from monday\\ to thursday. How many computers are now in the server room?\\
    Q[3]: Michael had 58 golf balls. On tuesday, he lost 23 golf balls. On wednesday, he lost 2 more. How many golf\\ balls did he have at the end of wednesday?\\
    Q[4]: Olivia has \$23. She bought five bagels for \$3 each. How much money does she have left?\\
    \textcolor{brown}{Assistant}\\
    A[1]: Shawn started with 5 toys. If he got 2 toys each from his mom and dad, then that is 4 more toys. 5 + 4 = 9. \\The answer is 9.\\
    A[2]: There were originally 9 computers. For each of 4 days, 5 more computers were added. So 5 * 4 = 20 \\computers were added. 9 + 20 is 29. The answer is 29.\\
    A[3]: Michael started with 58 golf balls. After losing 23 on tuesday, he had 58 - 23 = 35. After losing 2 more, he had \\35 - 2 = 33 golf balls. The answer is 33.\\
    A[4]: Olivia had 23 dollars. 5 bagels for 3 dollars each will be 5 x 3 = 15 dollars. So she has 23 - 15 dollars left. \\23 - 15 is 8. The answer is 8.\\
    \textcolor{violet}{User}\\
    Q[1]: A garden produced 237 potatoes, 60 fewer cucumbers and twice as many peppers than the cucumbers. How \\many vegetables did the garden produce?\\
    Q[2]: John's cow weighs 400 pounds. It increased its weight to 1.5 times its starting weight. He is able to sell the cow\\ for \$3 per pound. How much more is it worth after gaining the weight?\\
    Q[3]: John writes 20 pages a day. How long will it take him to write 3 books that are 400 pages each?\\
    Q[4]: James has a rainwater collection barrel. For each inch of rain he collects 15 gallons. On Monday it rained 4 inches\\ and on Tuesday it rained 3 inches. He can sell water for \$1.2 per gallon. How much money did he make from selling\\ all the water?\\
    \textcolor{brown}{Assistant}\\
    A[1]: The garden produced 237 - 60 = 177 cucumbers. The garden produced 177 * 2 = 354 peppers. The garden \\produced 237 + 177 + 354 = 768 vegetables. The answer is 768.\\
    A[2]: The cow initially weighs 400 * 1.5 = 600 pounds. So it gained 600 - 400 = 200 pounds. It is worth 200 * 3 = 600 \\dollars more. The answer is 600.\\
    A[3]: He wants to write 3 * 400 = 1200 pages. So it will take him 1200 / 20= 60 days The answer is 60.\\
    A[4]: It rained 3 + 4 = 7 inches So he collected 7 * 15 = 105 gallons So he makes 105 * 1.2 = 126 from selling the \\water. The answer is 126.\\
    \textcolor{violet}{User}\\
    \{four test questions\}\\
    \textcolor{brown}{Assistant}\\
    \{four test answers.\}\\
    \bottomrule[1.2pt]
    \end{tabular} 
}
\caption{An example GPT-3.5~(ChatGPT) and GPT-4 prompt we use for batch prompting. Specifically, the task instruction is given in the system message. In the next a few rounds, one batch of in-context exemplars is input in one round as the role ``user", and the answers are output as the role ``assistant". In the final round, test samples' contexts are input and the model outputs the answers.}
\label{tab:prompt chatgpt}
\end{table*}

\end{document}